\documentclass{article}

\usepackage{arxiv}

\usepackage[utf8]{inputenc} % allow utf-8 input
\usepackage[T1]{fontenc}    % use 8-bit T1 fonts
\usepackage{hyperref}       % hyperlinks
\usepackage{url}            % simple URL typesetting
\usepackage{booktabs}       % professional-quality tables
\usepackage{amsfonts}       % blackboard math symbols
\usepackage{nicefrac}       % compact symbols for 1/2, etc.
\usepackage{microtype}      % microtypography
\usepackage{bm} 
\usepackage{amsmath, txfonts} 
\usepackage{multirow}
\usepackage{here}
\usepackage[pdftex]{graphicx}

\def\i<#1>{\langle #1 \rangle}   % inner product
\def\g(#1){g(#1;\theta_k)}

\def\mbl{\mathcal{X}_{\mathrm{L}} ^ \mathrm{mb} }
\def\mbul{\mathcal{X}_{\mathrm{UL}} ^ \mathrm{mb} }

\def\mblaug{\mathcal{X}_{\mathrm{Laug}} ^ \mathrm{mb} }
\def\mbulaug{\mathcal{X}_{\mathrm{ULaug}} ^ \mathrm{mb} }

\def\dl{\mathcal{D}_\mathrm{L}}
\def\dul{\mathcal{D}_\mathrm{UL}}

\newcommand{\argmax}{\mathop{\rm arg~max}\limits}

\title{Gradient-based Data Augmentation for Semi-Supervised Learning}

\date{} 					% Or removing it

\author{
  Hiroshi Kaizuka  $^1$ \\
  NS Solutions Corporation\\
}

\begin{document}

\maketitle

\begin{abstract}
In semi-supervised learning (SSL), a technique called \textit{consistency regularization (CR)} achieves high performance. It has been proved that the diversity of data used in CR is extremely important to obtain a model with high discrimination performance by CR. We propose a new data augmentation (\textit{Gradient-based Data Augmentation (GDA)}) that is deterministically calculated from the image pixel value gradient of the posterior probability distribution that is the model output. We aim to secure effective data diversity for CR by utilizing three types of GDA. On the other hand, it has been demonstrated that the mixup method for labeled data and unlabeled data is also effective in SSL. We propose an SSL method named  \textit{MixGDA} by combining various mixup methods and GDA. The discrimination performance achieved by MixGDA is evaluated against the 13-layer CNN that is used as standard in SSL research. As a result, for CIFAR-10 (4000 labels), MixGDA achieves the same level of performance as the best performance ever achieved. For SVHN (250 labels, 500 labels and 1000 labels) and CIFAR-100 (10000 labels), MixGDA achieves state-of-the-art performance.
\\
\\
\end{abstract}

\textbf{\large Notation}\\
In this paper, the following notation is used for $n$-dimensional real vectors $x$ and $y \in \mathbb{R}^n $ and $n$-dimensional probability distributions $p$ and $q$.

\begin{table}[h]
  \centering
\begin{tabular}{ll}
	$ x_{\_i} $                         										&  Element $i$ of $x$   \\
	$ \| x \|_p=(|x_{\_1}|^p+|x_{\_2}|^p+ \cdots +|x_{\_n}|^p)^{1/p} $  		&  $L_p$  norm of $x$   \\
	$ \i<x,y>=x_{\_1}y_{\_1} + x_{\_2}y_{\_2} +  \cdots +  x_{\_n}y_{\_n} $	&  Inner product of $x$ and $y$   \\
	$ x \odot y = [x_{\_1}y_{\_1}   \cdots  x_{\_n}y_{\_n}] $                 	&  Element-wise (Hadamard)  product of $x$ and $y$   \\
	$ \mathrm{cos}(x,y) = \frac{\i<x,y>}{\| x \|_2\| y \|_2} $				&  Cosine of the angle between x and y \\
	$ H(p)=-\sum^{n}_{i=1}p_{\_i}{\log}p_{\_i} $								&  Shannon entropy of $p$   \\
     $ \mathrm{CE}(p,q) =-\sum^{n}_{i=1}p_{\_i}{\log}q_{\_i} $				&  Cross-entropy of $p$ and $q$   \\
     $ D_{\mathrm{KL}}(p\|q) = \mathrm{CE}(p,q) - H(p) $ 				&  Kullback-Leibler divergence of $p$ and $q$   \\
\end{tabular}
\end{table}

% keywords can be removed
%\keywords{First keyword \and Second keyword \and More}

\footnotetext[1]{\: The current e-mail address is hiro.kaizuka0224@gmail.com.}

%%%%%%%%%%%%% Section 1 %%%%%%%%%%%%%%%%%%%%%%%%%%

\section{Introduction}\label{sec1}

When solving the problem of classifying images into $K$ classes by convolutional neural networks (CNNs), images related to the tasks are collected as training data. Manually assigning a label (indicating the class to which the image belongs) for each image is a time consuming task when the number of training data is large. Therefore, a situation may occur where the number of unlabeled data is much larger than the number of labeled data. If the number of labeled data is not sufficient, supervised learning (SL) using only labeled data can not achieve high generalization performance. Therefore, it is expected to improve the generalization performance of CNNs by utilizing the unlabeled data existing in a large amount. Semi-supervised learning (SSL) is a method to realize such expectation.\\
\\[-2mm]
Various methods have been proposed for SSL. Among them, there is a group of methods called \textit{consistency regularization (CR)}. In recent years, the best results for SSL have been realized by CR. CR defines two functions $f_\mathrm{target} (u;\theta)$ and $f(u;\theta)$ determined from an unlabeled data sample $u$ and CNN (weights $\theta$, posterior probability distribution $g(u;\theta)$) . These are functions that can be assumed to be natural that the difference between the two is small. CR optimizes $\theta$ so that the sum of the loss function indicating the difference between $f_\mathrm{target} (u;\theta)$ and $f(u;\theta)$ and the loss function for labeled data samples is minimized.\\
\\[-2mm]
For an unlabeled data sample $u$, there is a CR method where $f_\mathrm{target} (u;\theta) = \g(u)$ and $f(u;\theta) = g(\mathrm{Augment} (u);\theta_k)$. Here, $\mathrm{Augment} (u)$ is generated by performing image conversion on $u$ within a range that does not hinder image discrimination. Since this method is a simple and straightforward method, if excellent performance can be achieved with this method, it should be a highly versatile SSL method. Unsupervised Data Augmentation (UDA) \cite{26, 27} has succeeded in significantly improving the performance of SSL by a method pursuing this idea thoroughly. UDA is a definitive result that proves that the diversity of $\mathrm{Augment} (u)$ is extremely effective for improving the accuracy of SSL. UDA \cite{26} uses AutoAugment \cite{4} to generate various $\mathrm{Augment} (u)$. However, AutoAugment needed to determine the optimal data augmentation for each dataset by performing a computationally intensive search. In addition, the constructed optimal data augmentation itself was an image transformation with a high computational load. RandAugment \cite{5} is a data augmentation method that greatly reduces these disadvantages of AutoAugment. UDA \cite{27} (called UDA\_RA) realizes the diversity of $\mathrm{Augment} (u)$ using RandAugment. In practice, UDA\_RA prepares $\mathrm{Aug} (u) = \{ \text{100 images obtained by applying 100 different image conversions to $u$} \}$ in advance for each unlabeled data sample $u$ to reduce training time. UDA\_RA generates $\mathrm{Augment} (u)$ by randomly selecting data samples from $\mathrm{Aug} (u)$ for each unlabeled data sample $u$ included in the minibatch of each training step. The data sample included in $\mathrm{Aug} (u)$ is $\mathrm{Cutout} (\mathrm{policy_2} (\mathrm{policy}_1 (u)))$ (Cutout \cite{6} ) using two kinds of policies, $\mathrm{policy}_1$ and $\mathrm{policy}_2$, randomly selected from
\begin{equation*}
	\left\{
		\text{policy} = 
			\begin{pmatrix}
				\text{transform} \\ \text{probability} = 0.5 \\ \text{magnitude}
			\end{pmatrix}
			\: \middle| \:
			\begin{matrix}
				\text{transform} \in
					\left\{ 
					\begin{matrix}
						\text{Invert,Cutout,Sharpness,AutoContrast,}\\
						\text{Posterize,ShearX,TranslateX,TranslateY,}\\
						\text{ShearY,Rotate,Equalize,Contrast,Color,}\\
						\text{Solarize,Brightness}
					\end{matrix}
					\right\} \\
				\text{magnitude} \in \{1,2, \cdots ,9\}	
			\end{matrix}					
	\right\}.
\end{equation*}
UDA\_RA realizes the diversity of $\mathrm{Augment} (u)$ by using many kinds of image transformations at random. There, the characteristic information of $u$ that appears as the training progresses is not used at all. As a result, UDA\_RA requires a lot of training. However, the obtained discrimination performance is extremely high. For example, in UDA\_RA for CIFAR-10 using Wide-ResNet-28-2 \cite{28}, one minibatch consists of 64 labeled data and 320 unlabeled data. The training is then repeated 400k times, achieving the world's highest performance for multiple datasets. Thus, UDA\_RA can be called a brute force technique in a sense. The motivation of this research is to answer the following questions: \\
\\[-2mm]
Is it possible to construct effective data augmentation for SSL by effectively using the posterior probability distribution $\g (u)$ obtained in the $k$-th training?

%%%%%%%%%%%%% Section 2 %%%%%%%%%%%%%%%%%%%%%%%%%%
\section{Overview of our method (\textit{MixGDA})}\label{sec2}
This paper assumes that unlabeled data samples can always be classified into one class, as in previous SSL research. Figure 1 shows the overall structure of our method (called \textit{MixGDA}).

\newpage
\begin{figure}[H]
  \centering
  \includegraphics[keepaspectratio, width=\linewidth]{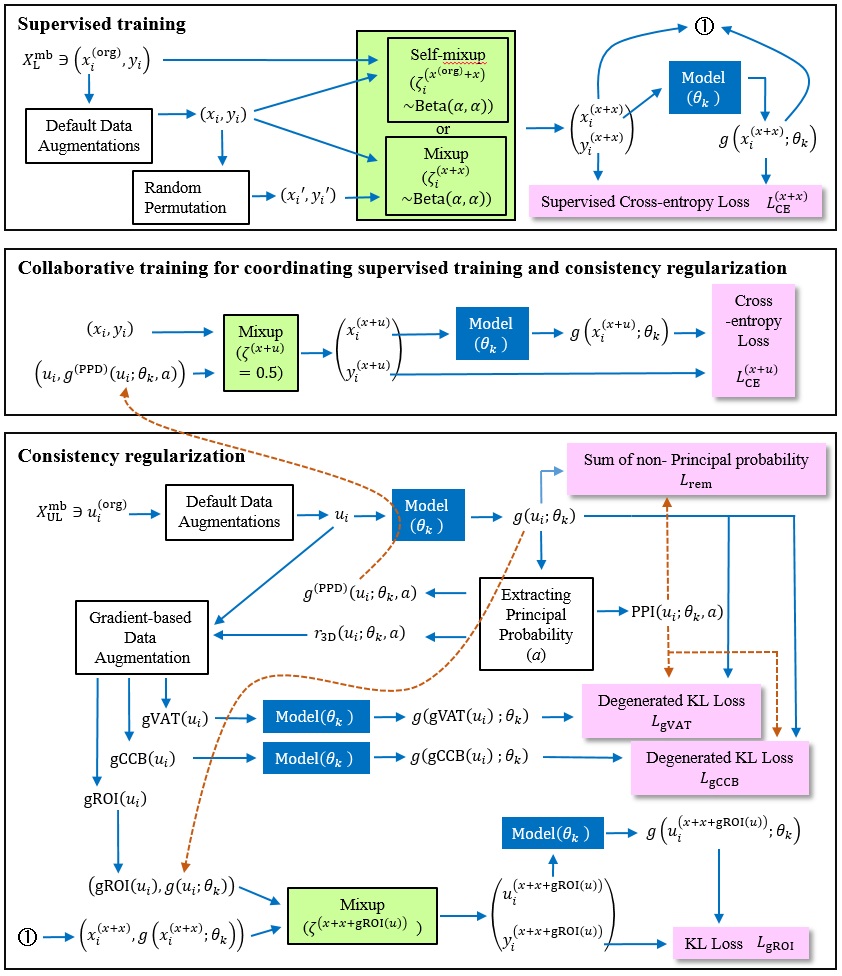}
  \caption{This figure shows the overall structure of \textit{MixGDA} except for the \textit{Inner} that realizes aggregation and separation. The loss function to be minimized is the weighted sum of the six types of loss functions shown in the figure. Here, the default data augmentations include stochastic processing. However, all processes except Mixup and Self-mixup used for supervised training are computed deterministically. Whether to use Mixup or Self-mixup in supervised training is selected for each data set.}
  \label{fig1}
\end{figure}
\newpage

\paragraph{Gradient-based data augmentation (\textit{GDA}).}
One naive idea is to exploit the $u$-gradient of the Shannon entropy of $\g (u)$. However, the important information for discrimination is the probability of having a higher value in  $\g (u)$. Unfortunately, such implications for discrimination are not included in Shannon entropy. Therefore, we find the maximum probability in  $\g (u)$ as $g_\mathrm{max} (u; \theta_k)$. Next, we calculate the sum of the probabilities of values smaller than $a \times g_\mathrm{max} (u; \theta_k) (0 \leq a \leq 1)$, and define it as $g_\mathrm{rem} (u; \theta_k,a)$. Then, we consider a probability distribution consisting of $g_\mathrm{rem} (u; \theta_k,a)$ and probabilities not included in $g_\mathrm{rem} (u; \theta_k,a)$. We evaluate the Shannon entropy of this degenerated probability distribution as degenerated entropy. For example, the degenerated entropy is calculated as follows:
\begin{equation*}
	\g(u) =
		\begin{bmatrix}
			0.01 \\ 0.01 \\ 0.01 \\ 0.01 \\ 0.01 \\  0.01 \\ 0.04 \\ 0.1 \\ 0.3 \\ 0.5
		\end{bmatrix}
		\xrightarrow[a = 0.3]{}
		\begin{bmatrix}
			g_\mathrm{rem} (u; \theta_k,a) = 0.2 \\ 0.3 \\ 0.5 
		\end{bmatrix},
\end{equation*}
\begin{equation*}
	\text{Degenerated Entropy}(u; \theta_k,a) = -0.5 \log 0.5 - 0.3 \log 0.3  - 0.2 \log 0.2.
\end{equation*}
The degenerated entropy is complex information of the randomness of the probability distribution and the probability contributing to discrimination. `$a$' is one of the important hyperparameters. This paper proposes three new types of data augmentation $\mathrm{gVAT} (u)$, $\mathrm{gCCB} (u)$, and $\mathrm{gROI} (u)$ effective for SSL. These are computed deterministically using the $u$-gradient of the degenerate entropy. $\mathrm{gVAT} (u)$ and  $\mathrm{gCCB} (u)$ belong to the adversarial data augmentation. $\mathrm{gROI} (u)$ holds the pixel value of $u$ in the region of interest (\textit{ROI}). However, in regions other than ROI, the contrast of  $\mathrm{gROI} (u)$ is smaller than the contrast of $u$. Therefore,  $\mathrm{gROI} (u)$ is a data sample in which the information contained in the ROI of $u$ is emphasized. In this sense,  $\mathrm{gROI} (u)$ is a collaborative data augmentation that is opposite to adversarial data augmentation.

\paragraph{Mixup.}
Mixup \cite{29}, which is effective for improving the performance of supervised learning, is also applied to SSL and contributes to the improvement of discrimination accuracy. Mixup requires label data in addition to image data. Therefore, in order to apply Mixup to an unlabeled data sample $u$, it is necessary to artificially generate a label (\textit{fake label}) for $u$. Various fake labels calculated from $\g(u)$ are often used. When Mixup is applied to both labeled data and unlabeled data, various variations are possible. For simplicity of description, let $\mbl$ and $\mbul$ be a minibatch composed of labeled data and a minibatch composed of unlabeled data, respectively. Interpolation Consistency Training (ICT) \cite{25} uses Mixup for data samples included in $\mbl$ and Mixup for data samples included in $\mbul$. In MixMatch \cite{2}, Mixup is applied to the data samples included in $\mbl \cup \mbul$. As a result, labeled data and unlabeled data may be mixed up. MixGDA uses three types of Mixups:\\
\\[-2mm]
\begin{tabular}{cl}
	1)  &Mixup of $x \in \mbl$ and $x' \in \mbul$. The resulting data is denoted as $\mathrm{Mix}_\lambda (x, x ')$ (the mixing ratio $\lambda \sim \mathrm{Beta} (\alpha, \alpha)$),\\
	2)  &Mixup of $\mathrm{Mix}_\lambda (x, x ')$ and $\mathrm{gROI} (u)$ (the mixing ratio is fixed),\\
	3)  &Mixup of $x \in \mbl$ and $u \in \mbul$ (the mixing ratio = 0.5).\\
\end{tabular}
1) contributes to the improvement of the discrimination accuracy achieved by supervised training, and 2) contributes to the improvement of the discrimination accuracy achieved by unsupervised training. On the other hand, 3) enhances the synergistic effect of supervised training and unsupervised training and improves the achievable discrimination accuracy. In our experiments, using 1) for SVHN, CIFAR-10 (labeled data = 250) and CIFAR-100 reduced the discrimination accuracy achievable. The reason for this is presumed that the mixed up data is not valid as training data. In the case of SVHN, data generated by mixing up data belonging to different classes (for example, 3 and 8) may be unnatural as training data. On the other hand, in the case of CIFAR-10 (labeled data = 250) and CIFAR-100, the achievable discrimination accuracy is low. Therefore, training data generated by mixing up data belonging to different classes cannot contribute to the realization of \textit{the low-density separation assumption}.\\
\\[-2mm]
In our experiment, $\mathrm{Augment} (x)$ is generated by applying default data augmentation (random translation, random horizontal flipping) to a labeled data sample $x$. Then, supervised learning is performed using ($\mathrm{Augment} (x)$, the label of $x$). From the above discussion, mixing up $x$ with $\mathrm{Augment} (x)$ should produce gentle training data. We call this method \textit{Self-mixup}. Since the label of the self-mixed data is the label of x itself, the data generated by Self-mixup exists near the decision boundary. As a result, data generated by Self-mixup is more likely to retain validity as training data than data generated by normal Mixup.

\paragraph{Aggregation and separation (\textit{Inner}).}
Let $\angle (\g (u_1),\g (u_2))$ be the angle between $\g (u_1)$ and $\g (u_2)$ for two unlabeled data samples $u_1$ and $u_2$. Then, as the number of training iterations $k$ increases, the following relation should be satisfied:
\begin{equation*}
	u_1 \text{ and } u_2 \text{ belong to the same class.} \Rightarrow  \angle (\g (u_1),\g (u_2)) \rightarrow 0 
\end{equation*}
\begin{equation*}
	u_1 \text{ and } u_2 \text{ belong to different classes.} \Rightarrow  \angle (\g (u_1),\g (u_2)) \rightarrow \pi /2 
\end{equation*}
Therefore, we introduce a method that minimizes $1-\i<\g (u_1),\g (u_2)>$ if $\angle (\g (u_1),\g (u_2)) \leq \pi /6$ and minimizes $\i<\g (u_1),\g (u_2)>$ if $\angle (\g (u_1),\g (u_2)) \geq \pi /3$. We also introduce a method to minimize $g_\mathrm{rem} (u; \theta_k, a)$.

\paragraph{Reliability function.}  
The loss function of a data sample $z$ is a non-negative loss $l (\g (z), g (\mathrm{Augmen}t (z); \theta_k + \theta))$ representing the difference between $\g (z)$ and $g (\mathrm{Augmen}t (z); \theta_k + \theta)$. The weight $\theta$ is updated by the update formula $\theta_{k+1} = \theta_k - l_r \cdot \nabla_\theta l (\g (z), g (\mathrm{Augmen}t (z); \theta_k + \theta)) |_{\theta = 0}$. However, in the first half of the training, the reliability of $\g (z)$ is not enough. Therefore, it is necessary to evaluate the reliability of $\g (z)$ and multiply the degree of reliability by $l (\g (z), g (\mathrm{Augmen}t (z); \theta_k + \theta))$. We construct various reliability functions based on the following fact 1.\\
\\[-2mm]
\textbf{Fact 1}\\
\textit{The following relationships hold for arbitrary K-dimensional probability distributions p and q.}
\begin{equation}
	0 \leq \frac{H(p)}{\log K} \leq 1  \text{, and $0 = \frac{H(p)}{\log K}$  iff $p$ is one-hot.}
\end{equation}
\begin{equation}
	\frac{1}{\sqrt{K}} \leq \|p\|_2 \leq 1\text{, and  $\|p\|_2 = 1$ iff $p$ is one-hot.}
\end{equation}
\begin{equation}
	0 \leq \i<p,q> \leq 1 \text{, and $\i<p,q> = 1$ iff $p=q$ and $p$ is one-hot.}
\end{equation}
\begin{equation}
	0 \leq \cos (p,q) \leq 1 \text{, and $\cos (p,q) = 1$ iff $p=q$.}
\end{equation}

\textbf{Proof:}\\
(1) is a famous inequality.

(Proof of (2))\\
Since $\sum_{i=1}^{K}p_i = 1$, then 
\begin{equation*}
	\sum_{i=1}^{K}p_i^2 - \frac{1}{K} = \sum_{i=1}^{K}p_i^2 - \frac{1}{K}\sum_{i=1}^{K}p_i  
	=  \sum_{i=1}^{K}\left\{ (p_i - \frac{1}{K})^2 + \frac{1}{K}p_i - \frac{1}{K^2} \right\}
	\geq \sum_{i=1}^{K}(\frac{1}{K}p_i - \frac{1}{K^2}) = \frac{1}{K}\sum_{i=1}^{K}p_i - \frac{1}{K} = 0
\end{equation*}
holds. Therefore, the inequality on the left-hand side of (2) holds. The sum of the elements of $p$ is 1 and all elements are nonnegative, so the following holds:
\begin{equation*}
	1= \left( \sum_{i = 1}^{K} p_i \right) ^2 = ||p||_2 + \sum_{i \neq j}p_ip_j \geq ||p||_2.
\end{equation*}
If $||p||_2 = 1$ holds, then
\begin{equation*}
	\sum_{i \neq j}p_ip_j = 0
\end{equation*}
must hold from the above equation. Therefore, if $p_h \neq 0$, then $p_i = 0 (\forall i \neq h)$. Therefore, $p$ is one-hot.

(Proof of (3))\\
Since the elements of $p$ and $q$ are all nonnegative, the inequality on the left-hand side clearly holds. Next, since
\begin{equation*}
\begin{split}
	\i<p,q> &= ||p||_2 ||q||_2 \cos(p,q) \leq ||p||_1 ||q||_1\cos(p,q) \;\; (\because 0 \leq p_i \leq 1 \text{, so } p_i^2 \leq p_i)  \\
	&= \cos(p,q) \;\; (\because ||p||_1= ||q||_1 = 1) 
\end{split}
\end{equation*}
holds, the inequality on the right-hand side of (3) holds. If $\i<p, q> = 1$, then $||p||_2 = ||q||_2 = \cos(p,q) = 1$. Therefore, from (2), $p$ and $q$ are both one-hot and have the same direction. Therefore, the condition of iff holds.

(Proof of (4))\\
If $\cos(p,q) = 1$, then  $\exists a \in \mathbb{R}; p = aq$. Therefore,
\begin{equation*}
	1 = \|p\|_1 = a||q||_1 = a \;\; (\because ||p||_1= ||q||_1 = 1) .
\end{equation*}
Thus, the condition of iff holds. \\
$\blacksquare$\\

For an unlabeled data sample $u$, we employ one of the following functions as the reliability function for $\g (u)$:
\begin{equation*}
	0 \leq 1 - \frac{H(\g(u))}{\log K} \leq 1 \;\; \text{ or } \;\; \frac{1}{\sqrt{K}} \leq \|\g(u) \|_2 \leq 1.
\end{equation*}
From (1) and (2) in Fact 1, these functions approach 1 as $\g(u)$ approaches one-hot. Here, for a labeled data sample $x$, the one-hot vector indicating the class to which $x$ belongs is denoted as $\mathrm{label} (x)$. For labeled data samples $x_i$ and $x_j$, we adopt
\begin{equation*}
	0 \leq \cos (\mathrm{Mix}_\lambda(\mathrm{label} (x_i), \mathrm{label} (x_j)), g(\mathrm{Mix}_\lambda(x_i, x_j); \theta_k)) \leq 1
\end{equation*}
as the reliability function for $g(\mathrm{Mix}_\lambda(x_i, x_j);\theta_k)$. When Self-mixup is applied to $x_i$, we adopt one of the following functions as the reliability function for $g(\mathrm{Mix}_\lambda(x_i, \mathrm{Augment}(x_i));\theta_k)$:
\begin{equation*}
	0 \leq \cos (\mathrm{label} (x_i), g(\mathrm{Mix}_\lambda(x_i, \mathrm{Augment}(x_i));\theta_k)) \leq 1
	\;\; \text{ or } \;\; 
	0 \leq \i<\mathrm{label} (x_i), g(\mathrm{Mix}_\lambda(x_i, \mathrm{Augment}(x_i));\theta_k)> \leq 1.
\end{equation*}

\paragraph{Training schedule and averaged model.}
We experiment using a 13-layer convolutional network architecture \cite{15} which is often used in SSL research. Many recent studies have used Wide-Resnet-28-2 \cite{28}. However, in this paper, no experiments were performed on Wide-Resnet-28-2. We apply Adam \cite{13} to the training. As shown in the learning schedule of Figure 2, the learning rate is reduced stepwise at regular intervals in the latter half of the training. In our experiments, the test error rates obtained after the training indicated by the brown circles showed a large variation. In this state, the inconvenience occurs that the discrimination accuracy is greatly affected by the timing at which the training is stopped. Therefore, we adopt the averaged model obtained by simply averaging the weights obtained after the training indicated by the brown circles. This is a device inspired by the work of SWA \cite{11,1}. At this point, the statistics (mean and standard deviation) of the batch normalization of the averaged model are undefined. We determine these statistics by injecting a number of minibatches consisting only of labeled data into the averaged model. The averaged model obtained in this way always exhibits higher discrimination performance than the model immediately after the end of training (see Table 2).

\begin{figure}[h]
  \centering
  \includegraphics[keepaspectratio, scale=0.4]{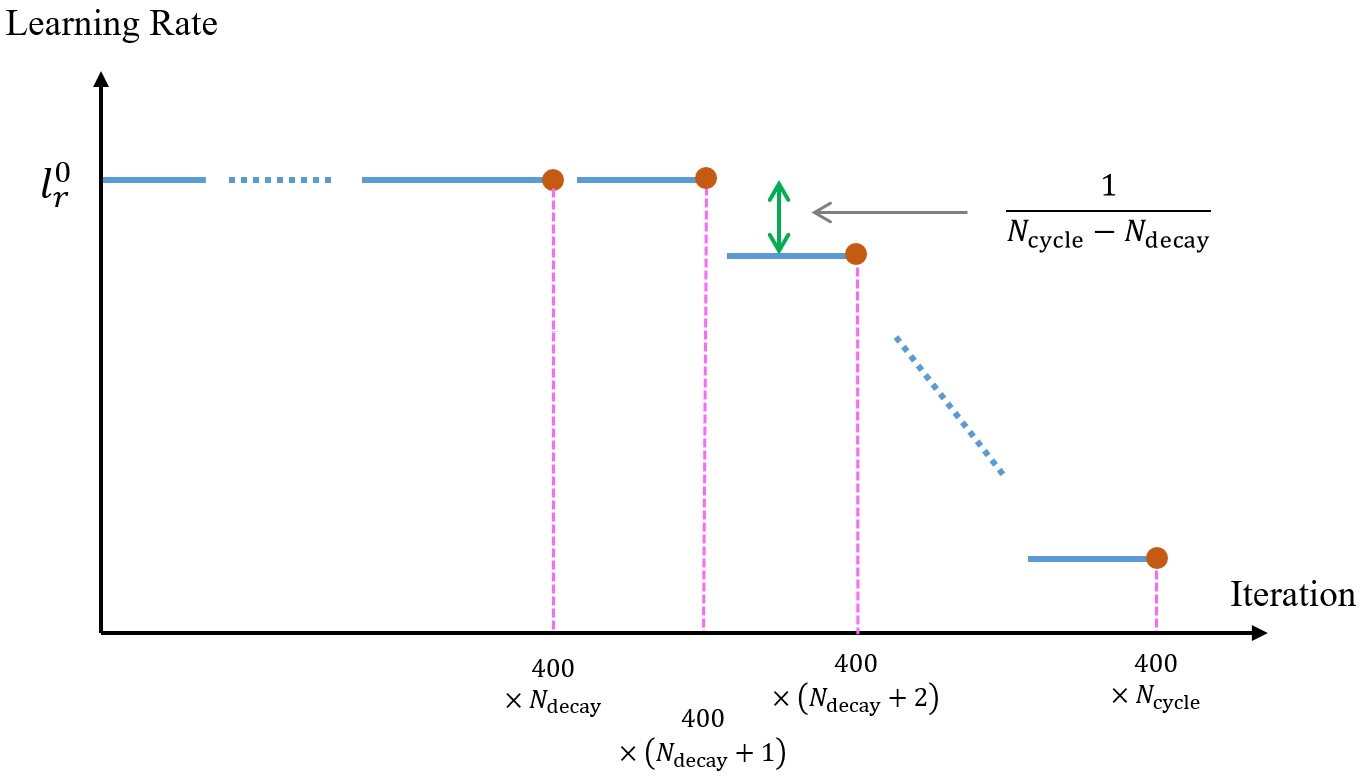}
  \caption{This figure shows Adam's learning rate schedule used in our experiments. The horizontal axis and the vertical axis indicate the number of iterations of minibatch update and the learning rate, respectively. $N_\mathrm{cycle}$ and $N_\mathrm{decay}$ are hyperparameters determined for each dataset. We obtain the weight of the averaged model by simply averaging the weight of the model obtained after the update calculation at the timing of the brown circle. This averaged model is the trained model that is actually used.}
  \label{fig2}
\end{figure}

%%%%%%%%%%%%% Section 3 %%%%%%%%%%%%%%%%%%%%%%%%%%
\section{Related Work}\label{sec3}
\paragraph{Degenerated entropy.}
The maximum probability in $\g (u)$ obtained when an unlabeled data sample $u$ is input to the model is denoted as $g_\mathrm{max}(u;\theta_k) = \max_{1 \leq j \leq K}g_j(u;\theta_k) $. $g_\mathrm{max}(u;\theta_k)$ is likely to be an expression of the characteristics of the class to which $u$ belongs. Therefore, we proposed \textit{ROI regularization (ROIreg)} that performs CR using $\nabla_r g_\mathrm{max}(u+r;\theta_k) |_{r=0}$ \cite{12}. However, $g_\mathrm{max}(u;\theta_k)$ may indicate a class to which $u$ does not belong. In particular, the frequency of such errors increases in the first half of training and for datasets with low discrimination accuracy. In such a case, the discrimination accuracy that ROIreg can achieve should decrease. We performed additional experiments with ROIreg using both VAT and entropy minimization for the case of reducing the number of labeled data and for CIFAR-100. Additional experiments were performed five times in each case using the model described in Table 6. The test error rates were $3.44 \pm 0.22$, $4.19 \pm 0.16$ and $12.22 \pm 5.65$ when the number of SVHN labeled data was 1000, 500 and 250, respectively. When the number of labeled data of CIFAR-10 was 4000, 2000 and 1000, the test error rates were $9.18 \pm 0.17$, $11.08 \pm 0.30$ and $15.12 \pm 0.48$, respectively. When the number of labeled data of CIFAR-100 was 10000, the test error rate was $37.89 \pm 0.38$. As expected, in the case of SVHN (250 labeled) and CIFAR-100 (10000 labeled), it can be seen that the achieved discrimination accuracy is poor. In this paper, we define the degenerated entropy by appropriately extracting the probabilities whose values are higher among the probabilities contained in $\g (u)$. We construct CR using the degenerated entropy instead of $g_\mathrm{max}(u;\theta_k)$.

\paragraph{GDA: gVAT.}
gVAT is based on the same concept as Virtual Adversarial Training (VAT) \cite{17}. VAT is a method inspired by Adversarial Training \cite{23}. For an unlabeled data sample $u$, VAT finds
\begin{equation*}
	r_0 = \argmax_{r \in \{r \mid \: \|r\|_2 = 1\}} D_\mathrm{KL}(\g(u) \mid\mid \g(u+r)),
\end{equation*}
sets a small positive number $\varepsilon$ as a hyperparameter, and updates $\theta$ to minimize $\mathrm{loss}(\g(u), g(u + \varepsilon r_0;\theta_k + \theta))$. Here, since $\nabla_r D_\mathrm{KL}(\g(u) \mid\mid \g(u+r))|_{r=0} = 0$, $r_0$ must be obtained as an eigenvector associated with the largest eigenvalue of the Hessian matrix of $D_\mathrm{KL}(\g(u) \mid\mid \g(u+r))$. However, the Hessian cannot be obtained by backpropagation because it is the information of second-order differentiation. For that purpose, VAT approximately calculates $r_0$ by performing one extra forward calculation and one  extra backward calculation. From this, it can be said that $u + \varepsilon r_0$ is a data augmentation that does not belong to GDA. $\mathrm{gVAT} (u)$ proposed in this paper is calculated using $\nabla_r \mathrm{Degenerated\  Entropy}(u + r; \theta_k, a)|_{r=0}$. This gradient can be determined by backpropagation using a computational graph that is also used for training. $\mathrm{gVAT} (u)$ can be said to be a first-order differentiation version of VAT.

\paragraph{GDA: gROI.}
When a new data sample $u$ is input to the trained model ($\theta =\theta_\mathrm{final}$), $y(u;\theta_\mathrm{final}) \in \mathbb{R}^K$ which is the input to the softmax and $g(u;\theta_\mathrm{final}) \in \mathbb{R}^K$ which is the output of the softmax are obtained. At this time, $u$ is determined to belong to the class corresponding to their maximum value. If we can show which region (\textit{the region of interest (ROI)}) in image $u$ is the basis for this decision, we can evaluate the validity of the model decision. For example, among the elements included in $\nabla_r y_\mathrm{max}(u + r; \theta_\mathrm{final})|_{r=0}$, elements whose absolute value is relatively large are extracted. There is a study interpreting that the pixels corresponding to the extracted elements constitute ROI \cite{22}. For such research, gROI targets the model in the $k$-th update calculation, not the trained model. The gROI focuses on $\mathrm{Degenerated\  Entropy}(u; \theta_k, a)$ instead of $g_\mathrm{max}(u;\theta_k)$, and generates $\mathrm{gROI} (u)$ using $\nabla_r \mathrm{Degenerated\  Entropy}(u + r; \theta_k, a)|_{r=0}$. In addition, $\g (u)$ obtained during training may not be reliable enough. Therefore, gROI also evaluates the reliability of $\g (u)$ and uses $\mathrm{gROI} (u)$ for CR.

\paragraph{GDA: gCCB.}
Data augmentation that randomly perturbs the contrast, color, and brightness of the entire image is used in UDA \cite{27}. gCCB divides an image into blocks, and perturbs the contrast, color, and brightness for each channel and each block. This perturbation is calculated deterministically. That is, using $\nabla_r \mathrm{Degenerated \ Entropy}(u+r;\theta_k,a) |_{r=0}$, the perturbation is determined so that the degenerated entropy changes most. Therefore, gCCB, like gVAT, is one of the adversarial data augmentations.

\paragraph{Self-mixup.}
The purpose of Self-mixup is to generate gentler training data than training data generated by normal Mixup. For this purpose, $\mathrm{Augment} (x)$ is generated by performing image conversion on a labeled data sample $x$ in a range that does not affect discrimination. Then, the data sample obtained by mixing up $x$ and $\mathrm{Augment} (x)$ is used as training data. The idea is the same as AugMix \cite{9}. AugMix generates two data samples from $x$, $\mathrm{AugmentAndMix}_1 (x)$ and $\mathrm{AugmentAndMix}_2 (x)$. These two data samples are constructed to be as diverse as possible. Then, Jensen-Shannon divergence of $\g (x)$, $g (\mathrm{AugmentAndMix}_1 (x) ; \theta_k)$ and  $g (\mathrm{AugmentAndMix}_2 (x) ; \theta_k)$ is adopted as the loss function. Therefore, it is possible to incorporate AugMix into MixGDA by performing supervised training using AugMix instead of Sefl-mixup and mixing $\mathrm{AugmentAndMix}_1 (x)$ into $\mathrm{gROI} (u)$. However, it is necessary to verify experimentally whether the discrimination performance achieved by MixGDA is improved.

\paragraph{Aggregation and separation.}
It is common in SSL research to assume that an unlabeled data sample always belongs to some class. Therefore, $\g (u)$ should be a one-hot vector. Since the Shannon entropy of a one-hot vector is 0, entropy minimization that minimizes the Shannon entropy of $\g (u)$ has been proposed \cite{8}. This method is combined with CR and contributes to the improvement of discrimination accuracy \cite{17,20}. However, since entropy minimization is an evaluation for a single unlabeled data sample, the variety of evaluations is limited. To increase the diversity of evaluation, it is effective to combine two unlabeled data samples for evaluation. From such an idea, we propose \textit{Inner}. Inner contains a term that promotes aggregation and a term that promotes separation. For two unlabeled data samples, if they can be expected to belong to the same class, the term promoting aggregation is applied to them. On the other hand, if they can be expected to belong to different classes, the term promoting separation is applied to them. The term that promotes separation does not appear to have been mentioned in the CR literature. However, in the field of similarity learning, aggregation and separation have been frequently used, as represented by contrastive loss \cite{3}. Therefore, the novelty of this paper is the Inner function form and the combination of CR with aggregation and separation.

%%%%%%%%%%%%% Section 4 %%%%%%%%%%%%%%%%%%%%%%%%%%
\section{Algorithm}\label{sec4}

In this paper, we consider a $K$-class classification problem for an image whose size is $N_r \times N_c$ and whose number of channels is $N_d$ ($N_d = 3$ for an RGB image). $\dl$ and $\dul$  denote a labeled data set and an unlabeled data set, respectively. For a labeled data sample $x \in \dl$, $\mathrm{label} (x)$ is a $K$-dimensional one-hot vector representing the class to which $x$ belongs. One minibatch consists of a set $\mbl$ consisting of $m_\mathrm{L}$ data samples randomly sampled from $\dl$  and a set $\mbul$ consisting of $m_\mathrm{UL}$ data samples randomly sampled from $\dl \cup \dul$.\\
\\[-2mm]
In this section, we assume a situation where the weight parameter of the model to be learned is $\theta=\theta_k$ after the $k$-th training finishes. For $u \in \dl \cup \dul $, execute a forward calculation
\begin{equation}
	\mathbb{R}^{N_{\mathrm{r}} \times N_{\mathrm{c}} \times N_{\mathrm{d}}} \ni u^{(\mathrm{org})} 
	\xrightarrow[\text{default data augmentation}]{}
	u
	\longrightarrow
	\text{Model\\ $(\theta = \theta_k)$}
	\longrightarrow 
	\g(u) \in  \mathbb{R}^K
	\notag
\end{equation}
using \textit{default data augmentation} (‘random horizontal flipping $\longrightarrow$ random translation’ or random translation) to obtain the posterior probability distribution $g(u;\theta_k)$.  The two types of data sets obtained by default data augmentation are described as follows:
\begin{equation}
	\mblaug 
          = \{ x = \text{default data augmentation}(x^\mathrm{(org)}) \mid x^\mathrm{(org)} \in \mbl \},	
\end{equation}
\begin{equation}
	\mbulaug
          = \{ u = \text{default data augmentation}(u^\mathrm{(org)}) \mid u^\mathrm{(org)} \in \mbul \}.	
\end{equation}
 Also, assume the natural assumption $m_\mathrm{L} \leq m_\mathrm{UL}$.

%%%%%%%%%%%%% Section 4.1 %%%%%%%%%%%%%%%%%%%%%%%%%%
\subsection{Overall structure of loss function}\label{sec4.1}
The overall structure of the loss function used in MixGDA is shown below:
\begin{equation}
\begin{aligned}
	\mathrm{Loss}(\theta)&=L_\mathrm{CE}^{(x+x)}(\theta)  &\text{(Supervised training (ST))}\\
	+\:&\delta_{\mathrm{gVAT}}L_{\mathrm{gVAT}}(\theta)+\rho_{\mathrm{gROI}} ( L_{\mathrm{gROI}}(\theta) + L_{\mathrm{rem}}(\theta) ) + \rho_{\mathrm{gCCB}}L_{\mathrm{gCCB}}(\theta) &\text{(Consistency regularization (CR))}  \\
	+\:& \delta^{(x+u)}L_{\mathrm{CE}}^{(x+u)}(\theta)  &\text{(collaborative training between ST and CR)}\\
	+\:& L_{\mathrm{inner}}(\theta) &\text{(Aggregation and separation)},
\end{aligned}
\end{equation}
where 
\begin{equation*}
\begin{aligned}
	&\delta_{\mathrm{gVAT}}, \delta^{(x+u)}=0 \text{\ or\ } 1 &\text{(On / off parameters)}, \\
	&\rho_{\mathrm{gROI}}, \rho_{\mathrm{gCCB}} \geq 0 &\text{(Hyperparameters)} .\\
\end{aligned}
\end{equation*}

%%%%%%%%%%%%% Section 4.2 %%%%%%%%%%%%%%%%%%%%%%%%%%
\subsection{Individual loss functions}\label{sec4.2}

\subsubsection{Gradient-based data augmentation (GDA)}\label{sec4.2.1}
In this section, for simplicity of description, it is assumed that a square RGB image ($N_d = 3$) with $N_r = N_c$ is the target and that the image can be divided into $M \times M$ size blocks without gaps. The datasets used in this paper are all $N_r = N_c = 32$, and $M = 4$ or $M = 8$ is used in the experiments. The following symbols are also used:
\begin{center}
	\begin{tabular}{ll}
		$1_ {M \times M}$ 			&$M \times M$ matrix where all elements are 1, \\
		$\Omega_q ^{M \times M}$	&$q$-th block when the image is divided into $M \times M$ size blocks. \\
	\end{tabular}
\end{center}

\paragraph{Degenerated entropy.}
In order to realize effective GDA for an unlabeled data sample $u$, it is important to extract appropriate scalar information from $K$-dimensional posterior probability distribution $\g (u)$. If such information can be extracted, the gradient information required for GDA can be calculated by backpropagation. Since the maximum probability
\begin{equation}
	g_{\mathrm{max}}(u;\theta_k)=\max_{1 \leq j \leq K} g_j (u;\theta_k)
\end{equation}
is likely to be the expression of the feature of the class to which $u$ belongs, it is a strong candidate for the scalar quantity to be found. However, during the first half or middle of the training, $u$ may belong to a class corresponding to a probability other than the maximum probability. Therefore, we set $m$ as a hyperparameter and pay attention to the top $m$ probabilities in $\g (u)$. Then the probability that $u$ belongs to any of the corresponding $m$ classes is much higher. However, in this case, it is necessary to calculate scalar information from these $m$ probabilities. Otherwise, the computation of the Hessian matrix that requires second-order differentiation would be required. Probabilities other than the top $m$ probabilities are considered noise. With this in mind, one honest idea comes to mind. That is, one probability (called the residual probability) consisting of the sum of probabilities other than the top $m$ probabilities is calculated. Then, an $m + 1$-dimensional probability distribution consisting of the top $m$ probabilities and the residual probability is formed, and its Shannon entropy is used as the scalar information to be obtained. The problem in this case is that $m$ should be reduced as the training progresses and the discrimination accuracy improves. Also, even when $m$ is reduced, the discontinuity of change cannot be wiped out because $m$ is a discrete value. Therefore, instead of ``the top $m$ probabilities'', we set $a \in [0,1]$ as the hyperparameter and adopt the probability group 
\begin{equation}
	\{  g_j (u;\theta_k) \mid g_j (u;\theta_k) \geq a \cdot g_{\mathrm{max}}(u;\theta_k), \: j=1,2, \cdots ,K \}.
\end{equation}
This makes it possible to fix $a$ to a constant value without changing a during training.

The above idea is realized as follows:

$\bullet$ \textit{Principal Probability Indicator:}
\begin{equation}
	\mathrm{PPI}(u;\theta_k,a) = \begin{bmatrix}
										\vdots \\
										\left\{
										\begin{array}{ll}
											1 & ( g_j (u;\theta_k) \geq a \cdot g_{\mathrm{max}}(u;\theta_k)) \\
											0 & (\text{else})
										\end{array}
										\right.
										\\
										\vdots
									\end{bmatrix}
									\begin{matrix}
										\  \\
										(j \\
										\ 
									\end{matrix}
	\in  \mathbb{R}^K.
\end{equation}
$\bullet$ \textit{Principal Probability Distribution:}
\begin{equation}
	g^{(\mathrm{PPD})}(u;\theta_k,a) = \frac{\mathrm{PPI}(u;\theta_k,a) \odot \g(u)}{\i<\mathrm{PPI}(u;\theta_k,a),\g(u)>}
	\in  \mathbb{R}^K.
\end{equation}
$\bullet$ \textit{Degenerated Entropy:}
\begin{equation}
	\mathrm{DE}(u;\theta_k,a) = H \left( \begin{bmatrix}
											\mathrm{PPI}(u;\theta_k,a) \odot \g(u) \\
											1-\i<\mathrm{PPI}(u;\theta_k,a),\g(u)>
										\end{bmatrix}
								\in  \mathbb{R}^{K+1}
								\right) ,
\end{equation}
\begin{equation*}
\left(\text{ where \ } 
	\begin{bmatrix}
		\mathrm{PPI}(u;\theta_k,a) \odot \g(u) \\
		1-\i<\mathrm{PPI}(u;\theta_k,a),\g(u)>
	\end{bmatrix}
\text{\ is a probability distribution.}
\right)
\end{equation*}
This degenerated entropy is the scalar information required.

The gradient required for GDA is calculated as
\begin{equation}
	r_\mathrm{3D} (u;\theta_k,a) = \nabla_r \mathrm{DE}(u+r;\theta_k,a)|_{r=0}
\end{equation}
using $\mathrm{DE}(u;\theta_k,a)$. The reliability of $\g (u)$ is evaluated by 
\begin{equation}
	d_\mathrm{rel}(\g(u)) = 1 - \frac{H(\g(u))}{\log K}.
\end{equation}

\paragraph{gVAT (adversarial data augmentation).}
We propose the following data augmentation using the gradient $r_\mathrm{3D} (u;\theta_k,a)$ and the hyperparameter $\varepsilon$:
\begin{equation}
	\mathrm{gVAT}(u) = u + \varepsilon \cdot \frac{r_\mathrm{3D} (u;\theta_k,a)}{\| r_\mathrm{3D} (u;\theta_k,a) \|_1}.
\end{equation}
This is a data augmentation based on the same concept as VAT. In other words, it is a data augmentation that perturbs $u$ in the direction that destroys the scalar information of interest most. gVAT, unlike VAT, does not require one extra forward propagation and one extra backward propagation. The loss function is
\begin{equation}
\begin{split}
	L_{\mathrm{gVAT}}(\theta) = &\frac{1}{m_\mathrm{UL}} \sum_{u \in \mbulaug}
	d_\mathrm{rel}(\g(u))   \\
	&\times D_{\mathrm{KL}} \left(
						\begin{bmatrix}
							\mathrm{PPI}(u;\theta_k,a) \odot \g(u) \\
							1-\i<\mathrm{PPI}(u;\theta_k,a),\g(u)>
						\end{bmatrix}
						\: \middle| \middle| \:
						\begin{bmatrix}
							\mathrm{PPI}(u;\theta_k,a) \odot g(\mathrm{gVAT}(u);\theta_k + \theta) \\
							1-\i<\mathrm{PPI}(u;\theta_k,a),g(\mathrm{gVAT}(u);\theta_k + \theta)>
						\end{bmatrix}
					\right).												
\end{split}
\end{equation}

\paragraph{gCCB (adversarial data augmentation).}
gVAT is a method of adversarially perturbing the pixel value for each pixel. gCCB is a method of adversarially perturbing contrast, color, and brightness. It would be more natural to apply these perturbations per image block rather than per pixel. gCCB multiplies the contrast by $1 + s_\mathrm{cont} (c, q)$ and simultaneously increases the brightness by $s_\mathrm{bri} (c, q)$ for the $c$ channel included in the image block $q$ of an unlabeled data $u$. The resulting image $\hat{u}$ can then be written as:
\begin{equation}
	\hat{u}(i,j,c) = (1 + s_\mathrm{cont}(c,q) )u(i,j,c) + s_\mathrm{bri}(c,q) 
	\; \; \;
	\left(
		(i,j) \in \Omega_q^{M_\mathrm{CCB} \times M_\mathrm{CCB}}, c=R,G,B
	\right).
\end{equation}
If we define the RBG images on $\Omega_q^{M_\mathrm{CCB} \times M_\mathrm{CCB}}$ of $u$ and $\hat{u}$ as 
\begin{equation*}
	u(q) =	\begin{bmatrix}
				u_R(q) \\
				u_G(q) \\
				u_B(q) \\
			\end{bmatrix} \in  \mathbb{R}^{M_\mathrm{CCB} \times M_\mathrm{CCB} \times 3}\;\text{ and }\;
	\hat{u}(q) =\begin{bmatrix}
					\hat{u}_R(q) \\
					\hat{u}_G(q) \\
					\hat{u}_B(q) \\
				\end{bmatrix} \in  \mathbb{R}^{M_\mathrm{CCB} \times M_\mathrm{CCB} \times 3} \;\;\;
	(q=1,2, \cdots ,Q),
\end{equation*}
respectively, $\hat{u}(q)$ can be expressed as
\begin{equation}
	\hat{u}(q) =\begin{bmatrix}
					(1 + s_\mathrm{cont}(R,q) )u_R(q) \\
					(1 + s_\mathrm{cont}(G,q) )u_G(q) \\
					(1 + s_\mathrm{cont}(B,q) )u_B(q) \\
				\end{bmatrix} +
				\begin{bmatrix}
					s_\mathrm{bri}(R,q)1_{M \times M} \\
					s_\mathrm{bri}(G,q)1_{M \times M} \\
					s_\mathrm{bri}(B,q) 1_{M \times M}\\
				\end{bmatrix} .
\end{equation}
Here, let
\begin{equation}
	r_\mathrm{3D} (q;u) =	
			\begin{bmatrix}
				S_R (q;u) \\
				S_G (q;u) \\
				S_B (q;u) 
			\end{bmatrix} \;\;\;
	(q=1,2, \cdots ,Q),
\end{equation}
the following approximate equation holds:
\begin{equation}
\begin{split}
	\mathrm{DE} \left( 	
					\begin{bmatrix}
						\vdots \\
						\hat{u}(q) \\
						\vdots
					\end{bmatrix} ;\theta_k,a
					\right)
	&=
	\mathrm{DE} \left( 
			\begin{bmatrix}
				\vdots \\	
				\begin{bmatrix}
					(1 + s_\mathrm{cont}(R,q) )u_R(q) \\
					(1 + s_\mathrm{cont}(G,q) )u_G(q) \\
					(1 + s_\mathrm{cont}(B,q) )u_B(q) \\
				\end{bmatrix} +
				\begin{bmatrix}
					s_\mathrm{bri}(R,q)1_{M \times M} \\
					s_\mathrm{bri}(G,q)1_{M \times M} \\
					s_\mathrm{bri}(B,q) 1_{M \times M}\\
				\end{bmatrix}  \\
				\vdots
			\end{bmatrix}   ;\theta_k,a\right) \\
	&\cong \mathrm{DE}(u;\theta_k,a) +
		\i<
			\begin{bmatrix}
				\vdots \\	
				\begin{bmatrix}
					S_R (q;u) \\
					S_G (q;u) \\
					S_B (q;u) 
				\end{bmatrix}  \\
				\vdots
			\end{bmatrix},
			\begin{bmatrix}
				\vdots \\	
				\begin{bmatrix}
					s_\mathrm{cont}(R,q) u_R(q) \\
					s_\mathrm{cont}(G,q) u_G(q) \\
					s_\mathrm{cont}(B,q) u_B(q) 
				\end{bmatrix} \\
				\vdots
			\end{bmatrix}  +
			\begin{bmatrix}
				\vdots \\	
				\begin{bmatrix}
					s_\mathrm{bri}(R,q)1_{M \times M} \\
					s_\mathrm{bri}(G,q)1_{M \times M} \\
					s_\mathrm{bri}(B,q) 1_{M \times M}
				\end{bmatrix}  \\
				\vdots
			\end{bmatrix} 
		>		\\
	&=  \mathrm{DE}(u;\theta_k,a)
	 + \sum_{q=1}^{Q} \sum_{c=R,G,B} \{ s_\mathrm{cont}(c,q) \cdot \i<S_c (q;u),u_c(q)>
										+ s_\mathrm{bri}(c,q) \cdot \i<S_c (q;u),1_{M \times M}> \}.
\end{split}
\end{equation}
In gCCB, a positive number $\mathrm{mag}_\mathrm{cont}$ and a positive number $\mathrm{mag}_\mathrm{bri}$ are set as hyperparameters to limit the perturbation to 
\begin{equation}
	s_\mathrm{cont}(c,q) = 										
		\left\{
		\begin{array}{ll}
			\mathrm{mag}_\mathrm{cont} \\
			\;\; \text{or}  \\
			-\mathrm{mag}_\mathrm{cont} 
		\end{array}
		\right. ,
	s_\mathrm{bri}(c,q) = 										
		\left\{
		\begin{array}{ll}
			\mathrm{mag}_\mathrm{bri} \\
			\;\; \text{or}  \\
			-\mathrm{mag}_\mathrm{bri} 
		\end{array}
		\right. .
\end{equation}
We adopt the strategy
\begin{equation}
\begin{matrix}									
	\left\{
	\begin{array}{ll}
		s_\mathrm{cont}(c,q) = \mathrm{mag}_\mathrm{cont} \times \mathrm{sign}(\i<S_c(q;u),u_c(q)>) \\
		s_\mathrm{bri}(c,q) = \mathrm{mag}_\mathrm{bri} \times \mathrm{sign}(\i<S_c(q;u),1_{M \times M}>)
	\end{array}
	\right. 
	& (q=1,2, \cdots, Q; \;\; c=R,G,B)
\end{matrix}
\end{equation}
from the idea of ``disturbing the image maximally = maximal increase of DE''. Since the pixel value of $u$ is normalized to $[-1, 1]$, 
\begin{equation}
	\mathrm{gCCB}(u)(i,j,c) = \mathrm{max}[ \mathrm{min} [ (1 + s_\mathrm{cont}(c,q))u(i,j,c) 
			+ s_\mathrm{bri}(c,q),1 ], -1 ]
\end{equation}
\begin{equation*}
	\left(
		(i,j) \in  \Omega_q^{M_\mathrm{CCB} \times M_\mathrm{CCB}},  c=R,G,B
	\right)
\end{equation*}
is applied to restrict $\mathrm{gCCB} (u)$ to $[-1, 1]$. The loss function is
\begin{equation}
\begin{split}
	L_{\mathrm{gCCB}}(\theta) = &\frac{1}{m_\mathrm{UL}} \sum_{u \in \mbulaug}
	d_\mathrm{rel}(\g(u))   \\
	&\times D_{\mathrm{KL}} \left(
						\begin{bmatrix}
							\mathrm{PPI}(u;\theta_k,a) \odot \g(u) \\
							1-\i<\mathrm{PPI}(u;\theta_k,a),\g(u)>
						\end{bmatrix}
						\middle| \middle|
						\begin{bmatrix}
							\mathrm{PPI}(u;\theta_k,a) \odot g(\mathrm{gCCB}(u);\theta_k + \theta) \\
							1-\i<\mathrm{PPI}(u;\theta_k,a),g(\mathrm{gCCB}(u);\theta_k + \theta)>
						\end{bmatrix}
					\right).												
\end{split}
\end{equation}

\paragraph{gROI (collaborative data augmentation).}
Calculate
\begin{equation}
	r_\mathrm{2D}(u,\Omega_q^{M_\mathrm{ROI} \times M_\mathrm{ROI}})
	= \frac{1}{\| r_\mathrm{3D} (u;\theta_k,a) \|_1}\sum_{(i,j) \in \Omega_q^{M_\mathrm{ROI} \times M_\mathrm{ROI}}}
	\sum_{c=R,G,B}| r_\mathrm{3D}(u;\theta_k,a)(i,j,c) |
\end{equation}
for each image block $q$. It is considered that image blocks with larger $r_\mathrm{2D}(u,\Omega_q^{M_\mathrm{ROI} \times M_\mathrm{ROI}})$ contain more important information for discrimination. Therefore, if $\lambda_\mathrm{rate} \in  [0,1]$ is set as a hyperparameter and the image area $\Omega_\mathrm{low}(u;\lambda_\mathrm{rate})$ is calculated as follows, $\Omega_\mathrm{low}(u;\lambda_\mathrm{rate})^\mathrm{c}$ can be considered to be the ROI of $u$:
\begin{equation}
	r_\mathrm{2D}(u,\Omega_{q_1}^{M_\mathrm{ROI} \times M_\mathrm{ROI}}) \leq \cdots \leq
	r_\mathrm{2D}(u,\Omega_{q_Q}^{M_\mathrm{ROI} \times M_\mathrm{ROI}})	,
\end{equation}
\begin{equation}
	\Omega_\mathrm{low}(u;\lambda_\mathrm{rate})
	= \left\{
		r_\mathrm{2D}(u,\Omega_{q_1}^{M_\mathrm{ROI} \times M_\mathrm{ROI}}), \cdots,
		r_\mathrm{2D}(u,\Omega_{q_l}^{M_\mathrm{ROI} \times M_\mathrm{ROI}}) \: \middle| \:
		\begin{matrix}
			\sum_{j=1}^{l-1}r_\mathrm{2D}(u,\Omega_{q_j}^{M_\mathrm{ROI} \times M_\mathrm{ROI}}) < \lambda_\mathrm{rate} \\
			\sum_{j=1}^{l}r_\mathrm{2D}(u,\Omega_{q_j}^{M_\mathrm{ROI} \times M_\mathrm{ROI}}) \geq \lambda_\mathrm{rate}
		\end{matrix}
	\right\}.
\end{equation}
By setting the hyperparameter $\zeta^{(x+x+\mathrm{gROI}(u))} \in (0.5,1]$, the following image $\mathrm{gROI}(u)$ is constructed:
\begin{equation}
	\mathrm{gROI}(u)(i,j,c) = 										
		\left\{
		\begin{array}{ll}
			u(i,j,c) & ((i,j) \notin \Omega_\mathrm{low}(u;\lambda_\mathrm{rate})) \\
			\frac{1- \zeta^{(x+x+\mathrm{gROI}(u))}}{\zeta^{(x+x+\mathrm{gROI}(u))}} \times u(i,j,c) 
				& ((i,j) \in \Omega_\mathrm{low}(u;\lambda_\mathrm{rate})) 			
		\end{array}.
		\right.
\end{equation}
$\mathrm{gROI}(u)$ is an image whose pixel values are suppressed except for ROI, and is the same image as $u$ in ROI. Therefore, the fake label of $\mathrm{gROI}(u)$ is considered to be $\g (u)$. The dataset
\begin{equation}
	\left\{
		(\mathrm{gROI}(u), \g(u))  \: \middle| \: u \in \mbulaug										
	\right\}
\end{equation}
is constructed according to this judgment.

\subsubsection{Mixup}\label{sec4.2.2}
MixGDA uses three types of mixups as shown below.
\paragraph{Supervised training (ST).}
A data sample $x_i \in \mblaug = \{ x_1, x_2, \cdots, x_{m_\mathrm{L}} \}$ has a label of $\mathrm{label} (x_i) = \mathrm{label} (x_i ^ {(\mathrm{org})})$. At this time, the mixed data $\{(x_i ^ {(x + x)}, y_i ^ {(x + x)})\}$ is generated as follows. This is a normal Mixup.
\begin{equation}
\begin{gathered}
	\zeta^{(x+x)}(x_i) \sim \mathrm{Beta}(\alpha, \alpha) , \\
	\left(
		\begin{pmatrix}
			x_1 \\
			\mathrm{label}(x_1)
		\end{pmatrix} , \cdots ,
		\begin{pmatrix}
			x_{m_\mathrm{L}} \\
			\mathrm{label}(x_{m_\mathrm{L}})
		\end{pmatrix}
	\right)
	\xrightarrow[\text{random permutation}]{}
	\left(
		\begin{pmatrix}
			x_1' \\
			\mathrm{label}(x_1')
		\end{pmatrix} , \cdots ,
		\begin{pmatrix}
			x_{m_\mathrm{L}}' \\
			\mathrm{label}(x_{m_\mathrm{L}}')
		\end{pmatrix}
	\right) \\
	\xrightarrow[\text{mixup}]{}
	\left\{
		\begin{pmatrix}
			x_i^{(x+x)} = \zeta^{(x+x)}(x_i)x_i + (1 - \zeta^{(x+x)}(x_i))x_i' \\
			y_i^{(x+x)} = \zeta^{(x+x)}(x_i)\mathrm{label}(x_i) + (1 - \zeta^{(x+x)}(x_i))\mathrm{label}(x_i')
		\end{pmatrix}	
		\;\;\;\; (i=1,2, \cdots ,m_\mathrm{L})	
	\right\}.
\end{gathered}
\end{equation}
We propose the following mixup that is gentler than the above mixup:
\begin{equation}
\begin{gathered}
	\zeta^{(x^{\mathrm{(org)}}+x)}(x_i) \sim \mathrm{Beta}(\alpha, \alpha),  \\
	\zeta^{(x^{\mathrm{(org)}}+x)}(x_i) = \max (\zeta^{(x^{\mathrm{(org)}}+x)}(x_i), 1 - \zeta^{(x^{\mathrm{(org)}}+x)}(x_i)), \\	
	\left\{
		\begin{pmatrix}
			x_i^{(x+x)} =\zeta^{(x^{\mathrm{(org)}}+x)}(x_i)x_i + (1 - \zeta^{(x^{\mathrm{(org)}}+x)}(x_i))x_i^{\mathrm{(org)}} \\
			y_i^{(x+x)} = \mathrm{label}(x_i) 
		\end{pmatrix}	
		\;\;\;\; (i=1,2, \cdots ,m_\mathrm{L})	
	\right\}.
\end{gathered}
\end{equation}
We call this mixup \textit{Self-mixup}. $x_i$ has greater data diversity than $x_i ^ {(\mathrm{org})}$. Therefore, in order to ensure the data diversity of $x_i ^ {(x + x)}$, the mixing ratio of $x_i$ should be larger than the mixing ratio of $x_i ^ {(\mathrm{org})}$. This is the reason why $\zeta^{(x^{\mathrm{(org)}}+x)}(x_i)$ is set to 0.5 or more by the above max operation. Whether to use normal mixup or Self-mixup should be decided for each dataset. Self-mixup may be appropriate for datasets where the discrimination target strongly depends on the geometrical shape or the discrimination accuracy achieved is low. Otherwise, just select the normal mixup. The loss function is the following function:
\begin{equation}
	L_\mathrm{CE}^{(x+x)}(\theta) = \frac{1}{m_\mathrm{L}} \sum_{i=1}^{m_\mathrm{L}}
					\mathrm{CE} \left(
						y_i^{(x+x)}, g(x_i^{(x+x)};\theta_k+\theta)
					\right).
\end{equation}

\paragraph{CR using gROI($u$).}
For a data sample $u_i \in \mbulaug = \{u_1, u_2, \cdots, u_ {m_\mathrm{UL}}\}$, $\mathrm{gROI}(u_i)$ is the same image as $u_i$ in the region important for discrimination. Therefore, it is reasonable to consider $\g (u_i)$ as a fake label of $\mathrm{gROI}(u_i)$. Produces the following mixed up data:
\begin{equation}
\begin{gathered}
	\begin{matrix}
		\mathrm{gROI}(u_1) & \cdots & \mathrm{gROI}(u_{m_\mathrm{L}}) 
				& \mathrm{gROI}(u_{m_\mathrm{L}+1}) & \cdots & \mathrm{gROI}(u_{m_\mathrm{UL}}) \\
		x_1^{(x+x)}              & \cdots & x_{m_\mathrm{L}}^{(x+x)}               
				& x_{m_\mathrm{L}+1}^{(x+x)} =x_1^{(x+x)}  & \cdots & x_{m_\mathrm{UL}}^{(x+x)} = x_{m_\mathrm{UL} -m_\mathrm{L} }^{(x+x)}
	\end{matrix}, \\
	0.5<\zeta^{(x+x+\mathrm{gROI}(u))} \leq 1 \text{, and $\zeta^{(x+x+\mathrm{gROI}(u))}$ is a fixed value that is constant for each dataset,} \\
	\left\{
		\begin{pmatrix}
			u_i^{(x+x+\mathrm{gROI}(u))} =\zeta^{(x+x+\mathrm{gROI}(u))}\mathrm{gROI}(u_i) 
					+ (1 - \zeta^{(x+x+\mathrm{gROI}(u))})x_i^{(x+x)} \\
			y_i^{(x+x+\mathrm{gROI}(u))} =\zeta^{(x+x+\mathrm{gROI}(u))}\g(u_i) 
					+ (1 - \zeta^{(x+x+\mathrm{gROI}(u))})y_i^{(x+x)}
		\end{pmatrix}	
		\;\;\;\; (i=1,2, \cdots ,m_\mathrm{UL})	
	\right\}.
\end{gathered}
\end{equation}
At a pixel $(m, n) \in \Omega_\mathrm{low}(u_i;\lambda_\mathrm{rate})$, 
\begin{equation*}
\begin{split}
	u_i^{(x+x+\mathrm{gROI}(u))}(m,n,c) &=\zeta^{(x+x+\mathrm{gROI}(u))} 
			\left\{ \frac{1- \zeta^{(x+x+\mathrm{gROI}(u))}}{\zeta^{(x+x+\mathrm{gROI}(u))}} \times u_i (m,n,c) \right\} 
			+ (1 - \zeta^{(x+x+\mathrm{gROI}(u))})x_i^{(x+x)}(m,n,c) \\
			&=  (1 - \zeta^{(x+x+\mathrm{gROI}(u))})  \{ u_i (m,n,c) + x_i^{(x+x)}(m,n,c) \}
\end{split}
\end{equation*}
holds. Therefore, in $(m, n) \in \Omega_\mathrm{low}(u_i;\lambda_\mathrm{rate})$, $u_i^{(x+x+\mathrm{gROI}(u))}$ is an image in which $u_i$ and $x_i^{(x+x)}$ are equally mixed and the contrast is reduced. The loss function is as follows:
\begin{equation}
	L_{\mathrm{gROI}}(\theta) = \frac{1}{m_\mathrm{UL}} \sum_{i=1}^{m_\mathrm{UL}}
	\frac{1}{2} \left\{ d_\mathrm{rel}(\g(u_i))  + d_\mathrm{rel}^{\mathrm{(label)}}(\g(x_i^{(x+x)}))  \right\}  \\
	\times D_{\mathrm{KL}} \left(
						y_i^{(x+x+\mathrm{gROI}(u))} \:\| \: g(u_i^{(x+x+\mathrm{gROI}(u))}; \theta_k + \theta)
					\right)												
\end{equation}
where
\begin{equation}
	d_\mathrm{rel}^{\mathrm{(label)}}(\g(x_i^{(x+x)})) = \left\{
		\begin{array}{ll}
			\cos (y_i^{(x+x)}, \g(x_i^{(x+x)})) 
				 &\text{if $x + x$ is the result of mixup.}    \\
			\cos (y_i^{(x+x)}, \g(x_i^{(x+x)}))  \;\; \text{or} \;\; \i<y_i^{(x+x)}, \g(x_i^{(x+x)})> 
				& \text{if $x + x$ is the result of Self-mixup.} 			
		\end{array}
		\right. 
\end{equation}
We apply Self-mixup to SVHN (the number of labeled data = 250, 500 and 1000), CIFAR-10 (the number of labeled data = 250) and CIFAR-100 (the number of labeled data = 10000).  As $d_\mathrm{rel}^{\mathrm{(label)}}(\g(x_i^{(x+x)}))$, $\i<y_i^{(x+x)}, \g(x_i^{(x+x)})>$ is used for SVHN, and $\cos (y_i^{(x+x)}, \g(x_i^{(x+x)}))$ is used for CIFAR-10 and CIFAR-100.

\paragraph{Collaborative training between ST and CR.}
The following upper and lower data samples are mixed up:
\begin{equation*}
\begin{matrix}
	x_1 & \cdots & x_{m_\mathrm{L}} \in \mblaug \\
	u_1 & \cdots & u_{m_\mathrm{L}} \in \mbulaug
\end{matrix}.
\end{equation*}
We use the sharper $g ^ {(\mathrm{PPD})}(u_i; \theta_k, a)$ defined by (11) instead of $\g (u_i)$ as the fake label of $u_i$. This is an idea inspired by the \textit{sharpening} used in MixMatch. And the mixed up data is generated as follows:
\begin{equation}
\begin{gathered}
	\zeta^{(x+u)} = 0.5, \\
	\left\{
		\begin{pmatrix}
			x_i^{(x+u)} =  \zeta^{(x+u)}x_i + (1 - \zeta^{(x+u)})u_i   \\
			y_i^{(x+u)} =  \zeta^{(x+u)} \mathrm{label}(x_i) + (1 - \zeta^{(x+u)}) g^{(\mathrm{PPD})}(u_i;\theta_k,a)
		\end{pmatrix} \;\;\;\
		(i=1,2, \cdots ,m_\mathrm{L})
	\right\}.
\end{gathered}
\end{equation}
The loss function is as follows:
\begin{equation}
	L_\mathrm{CE}^{(x+u)}(\theta) = \frac{1}{m_\mathrm{L}} \sum_{i=1}^{m_\mathrm{L}}
					\mathrm{CE} \left(
						y_i^{(x+u)}, g(x_i^{(x+u)};\theta_k+\theta)
					\right).
\end{equation}

\subsubsection{Aggregation and separation (Inner)}\label{sec4.2.3}
If two unlabeled data samples $u_i$ and $u_j$ are judged to belong to the same class, $\i<g(u_i;\theta_k + \theta), \g(u_j)> \rightarrow 1$ is promoted. This is a consequence of (3) described in Fact 1. If it is determined that they belong to different classes, the realization of $\i<g(u_i;\theta_k + \theta), \g(u_j)> \rightarrow 0$ should be promoted. Here, the cosine $\cos (\g(u_i), \g(u_j))$ of the angle formed by $\g (u_i)$ and $\g (u_j)$ as a $K$-dimensional vector is used to determine whether both belong to the same class or different classes. All elements of these two vectors are non-negative. Therefore, if $\cos (\g(u_i), \g(u_j)) \geq \cos (\pi/6)$, it is determined that they belong to the same class, and if $\cos (\g(u_i), \g(u_j)) \leq \cos (2 \pi/6)$, it is determined that they belong to different classes. Such a judgment is meaningless unless the reliability of $\g(u_i)$ and $\g(u_j)$ is high. Therefore, the above mechanism is realized between unlabeled data samples belonging to 
\begin{equation}
	\mbulaug(\beta)
          = \{ u \mid  \| \g(u) \|_2^2 \geq \beta^2,   u \in \mbulaug \}.	
\end{equation}
Here is the actual processing. For each $v_i \in \mbulaug(\beta)=[v_1, \cdots, v_n]$, $\mathrm{listW} (v_i)$ is obtained by random permutation of $[v_1, \cdots, v_n]$. Scan $\mathrm{listW} (v_i)$ from the top, and let $w_j$ that satisfies $\cos (\g(v_i), \g(w_j)) \leq \cos (2 \pi/6)$ for the first time be $u_\mathrm{diff} (v_i)$. If there is no such $w_j$, $u_\mathrm{diff} (v_i) = 0_K$. Also, $\mathrm{listW} (v_i)$ is scanned from the top, and $w_j$ that satisfies $\cos (\g(v_i), \g(w_j)) \geq \cos (\pi/6)$ for the first time is defined as $u_\mathrm{same} (v_i)$. If there is no such $w_j$, $u_\mathrm{same} (v_i) = 1_K$. Then, the following function is used as the loss function:
\begin{equation}
	L_\mathrm{inner}(\theta) = \frac{1}{m_\mathrm{UL}} \sum_{v \in \mbulaug(\beta)}
		\left\{
			\i<g(v;\theta_k + \theta), g(u_\mathrm{diff}(v), \theta_k)> + (1 - \i<g(v;\theta_k + \theta), g(u_\mathrm{same}(v), \theta_k)>)		
		\right\}.
\end{equation}
Note that $g(v;\theta_k + \theta)$ is a probability distribution, so that $\i<g(v;\theta_k + \theta), 1_K>=1$.

%%%%%%%%%%%%% Section 4.3 %%%%%%%%%%%%%%%%%%%%%%%%%%
\subsection{Training and averaged model}\label{sec4.3}

\paragraph{Training schedule.}
We employ training by Adam \cite{13} in all the experiments presented in this paper. We use the same training schedule used in \cite{17}. That is, in one cycle, the update by 400 minibatches is executed by $\mathrm{Adam} (l_r, \beta_1, \beta_2)$ using the same $l_r$, $\beta_1$ and $\beta_2$. The cycle starts at the 0-th cycle and ends at the $(N_\mathrm{cycle}-1)$-th cycle. Therefore, the total number of cycles is $N_\mathrm{cycle}$. In the cycle from 0 to $N_\mathrm{decay}$, $l_r = l_r ^ 0$, $\beta_1 = 0.9$ and $\beta_2 = 0.999$ are adopted.  For the $n_\mathrm{c}$-th cycle ($N_\mathrm{decay} + 1 \leq n_c \leq N_\mathrm{cycle}-1$) in the latter half of the training, $l_r = (N_\mathrm{cycle} - n_\mathrm{c}) / (N_\mathrm{cycle} - N_\mathrm{decay})$, $\beta_1 = 0.5$ and $\beta_2 = 0.999$ are adopted.

\paragraph{Averaged model.}
The weight of the model obtained after the $k$-th update is denoted as model($k$). The weight of the averaged model is calculated by the following formula:
\begin{equation}
	\mathrm{averaged\ model} = \frac{1}{N_\mathrm{cycle} - N_\mathrm{decay} +1} 
			\sum_{n_\mathrm{c} = N_\mathrm{decay}}^{N_\mathrm{cycle}} \mathrm{model}(400n_\mathrm{c}).
\end{equation}
Note that it is not necessary to store the weights of $N_\mathrm{cycle}-N_\mathrm{decay} + 1$ models for this calculation, and it is sufficient to calculate the variable weighted average sequentially.

\paragraph{Batch normalization statistics.}
In order to use the averaged model, the statistics (mean and standard deviation) of the batch normalization \cite{10} must be determined. We randomly sample 128 data from $\dl$ and construct one minibatch without applying default data enhancement. Using the 120 minibatches configured in this way, we determine the batch normalization statistics by performing 120 forward propagations. The update formula for the batch normalization statistic $\hat{x}_t$ is $\hat{x}_t = 0.9\hat{x}_{t-1} + 0.1x_t$. The same update formula is used for training using MixGDA.

%%%%%%%%%%%%% Section 5 %%%%%%%%%%%%%%%%%%%%%%%%%%
\section{Experiments}\label{sec5}
%

%%%%%%%%%%%%% Section 5.1 %%%%%%%%%%%%%%%%%%%%%%%%%%
\subsection{Comparison to Other Methods}\label{sec5.1}

\paragraph{Models.}
We perform the evaluation experiments of MixGDA using 13-layer CNN \cite{15} which is a standard model of SSL research. Details of the model are given in Table 6. In recent SSL research, evaluation experiments using Wide-ResNet-28-2 \cite{28} are increasing. However, in this paper, only the evaluation experiment for 13-layer CNN is performed. Evaluation experiments using Wide-ResNet-28-2 are for further study. MixMatch, which is the baseline for comparative evaluation, has been evaluated using Wide-ResNet-28-2. Evaluation experiments using 13-layer CNN have been performed only in a few cases. Therefore, in this paper, MixGDA and MixMatch are not sufficiently compared.

\paragraph{Datasets.}
We evaluate the performance of MixGDA against three standard benchmark datasets (SVHN \cite{18}, CIFAR-10 and CIFAR-100 \cite{14}). All of these datasets consist of $32 \times 32$ RGB images. SVHN is a close-up image of the house number and each image has a label corresponding to the number from 0 to 9 located at the center. CIFAR-10 and CIFAR-100 consist of natural images classified into 10 classes and 100 classes, respectively. In SVHN, the training set and the test set contain 73,257 images and 26,032 images, respectively. In CIFAR-10 and CIFAR-100, the training set and the test set contain 50,000 images and 10,000 images, respectively. The image data are normalized to $[-1, 1]$ by a simple linear transformation. In deep learning, ZCA normalization  \cite{14} is widely used as an extremely effective preprocessing for CIFAR-10. For example, ZCA normalization is used in comprehensive comparative evaluation of various SSLs  \cite{19} and ICT. However, if ZCA normalization is used in training, the image data samples input to the trained model must always be preprocessed by ZCA normalization. In addition, image data samples processed by ZCA normalization are very noisy and difficult to understand for humans. Therefore, ZCA normalization should not be used if possible. MixGDA is a method that does not use ZCA normalization. This is one of the features of MixGDA. It is not clear whether MixMatch uses ZCA normalization. As is common practice in SSL research, we construct a labeled data $\dl$ by randomly selecting a small number of data samples from the training set, and treat the remaining data samples as an unlabeled data $\dul$. For SVHN, only random translation by up to 2 pixels is used as default data augmentation. On the other hand, in the case of CIFAR-10 and CIFAR-100, both random horizontal flipping and random translation by up to 2 pixels are used.

\paragraph{Hyperparameters.}
In conventional SSL research, for example, even when the number of labeled data is 250, the hyperparameter is tuned using 1000 validation data. However, it should be considered how to divide 1250 labeled data into $\dl$ and validation data. Another important issue is how to reuse the validation data for training. These issues should be discussed properly as another theme. For this reason, we tune hyperparameters with test data. The values of the hyperparameters used in the experiment are summarized in Table 1.\\
\\[-2mm]
As can be seen from Table 1, the threshold ratio $a$ for judging the principal probability requires tuning for each dataset and each number of labeled data. This is a reflection of the fact that $a$ functions at the core of MixGDA. Most hyperparameters other than $a$ have a common value for each dataset. The guideline for setting $a$ is to set smaller values for datasets with lower discrimination accuracy. For the mixup used in supervised training, either normal mixup or self-mixup needs to be selected. The guideline for selection is to select self-mixup for simple images and datasets with low discrimination accuracy, and select mixup otherwise. We select mixup for CIFAR-10 (1000, 2000 and 4000 labels), otherwise select self-mixup.

\begin{table}[h]
  \centering
\begin{tabular}{c | c | c | c | c | c | c | c | c | c  }
\hline
\multicolumn{2}{c |}{Datasets} 		& \multicolumn{3}{| c |}{SVHN}				& \multicolumn{4}{| c |}{CIFAR-10}		& CIFAR-100	\\
\hline
\multicolumn{2}{c |}{Hyperparameters/Labels}				& 250		& 500		&1000		& 250		&1000		&2000		&4000			&10000 		\\ 	
\hline
\multirow{3}{*}{Training}		& $l_r^0$						& \multicolumn{3}{| c |}{0.001} 	& \multicolumn{4}{| c |}{0.00047}					& 0.00047		\\				
								& $N_\mathrm{cycle}$		& \multicolumn{3}{| c |}{120} 		& \multicolumn{4}{| c |}{500}						& 500			\\	
								& $N_\mathrm{decay}$		& \multicolumn{3}{| c |}{80}	 	& \multicolumn{4}{| c |}{460}						& 460			\\				
\hline
\multirow{2}{*}{Minibatch}		& $m_\mathrm{L}$				& \multicolumn{3}{| c |}{64} 		& \multicolumn{4}{| c |}{96}						& 96			\\				
								& $m_\mathrm{UL}$			& \multicolumn{3}{| c |}{96} 		& \multicolumn{4}{| c |}{96}						& 96			\\	
\hline
Collaborative					& $\delta^{(x+u)}$				& \multicolumn{3}{| c |}{0} 		& \multicolumn{4}{| c |}{1}							& 1				\\				
training						& $\zeta^{(x+u)}$				& \multicolumn{3}{| c |}{---} 		& \multicolumn{4}{| c |}{0.5}						& 0.5			\\
\hline
Supervised				     & Mixup						& \multicolumn{3}{| c |}{self} 		& self		& \multicolumn{3}{| c |}{mixup}		& self			\\	
\cline{2-10}			
training						& $\alpha$					& 0.1		& 0.2		& 0.3 		& \multicolumn{4}{| c |}{0.1}						& 0.1			\\
\hline
PPI								& $a$							& 0.1		& 0.5		& 0.5		& 0.1		& 0.2		& 0.35		& 0.4			& 0.2			\\
\hline
\multirow{2}{*}{gVAT}			& $\delta_\mathrm{gVAT}$	& \multicolumn{3}{| c |}{1} 		& \multicolumn{4}{| c |}{0}							& 0				\\				
								& $\varepsilon$				& \multicolumn{3}{| c |}{3.5} 		& \multicolumn{4}{| c |}{---}						& ---			\\	
\hline
\multirow{4}{*}{gCCB}			& $\rho_\mathrm{gCCB}$		& \multicolumn{3}{| c |}{1.2} 		& \multicolumn{4}{| c |}{2.0}						& 2.0			\\				
								& $M_\mathrm{CCB}$			& \multicolumn{3}{| c |}{8} 		& \multicolumn{4}{| c |}{8}							& 8				\\	
						& $\mathrm{mag}_\mathrm{cont}$	& \multicolumn{3}{| c |}{0.4} 		& \multicolumn{4}{| c |}{0.4}						& 0.4			\\
						& $\mathrm{mag}_\mathrm{bri}$		& \multicolumn{3}{| c |}{0.1} 		& \multicolumn{4}{| c |}{0.1}						& 0.2			\\
\hline
\multirow{4}{*}{gROI}			& $\rho_\mathrm{gROI}$		& \multicolumn{3}{| c |}{0.9} 		& \multicolumn{4}{| c |}{1.5}						& 1.5			\\				
								& $M_\mathrm{ROI}$			& \multicolumn{3}{| c |}{4} 		& \multicolumn{4}{| c |}{4}							& 4				\\	
				& $\lambda_\mathrm{rate}$					& \multicolumn{3}{| c |}{0.5} 		& \multicolumn{4}{| c |}{0.5}						& 0.5			\\
\cline{2-10}
				& $\zeta^{(x+x+\mathrm{gROI}(u))}$			& \multicolumn{3}{| c |}{0.8} 		& 0.8		& 0.75		& 0.8		&0.8			& 0.8			\\
\hline
Inner							& $\beta$						& \multicolumn{3}{| c |}{0.8} 		& \multicolumn{4}{| c |}{0.8}						& 0.5				\\	

\hline
\end{tabular}

\vspace{5mm}
  \caption{This table shows the values of the hyperparameters included in MixGDA. In the experiments in Table 2, the values in this table are used.}
  \label{tab1}
\end{table}

\paragraph{Results.}
Table 2 shows our experimental results. For SVHN, MixGDA achieves the highest performance. The reason for this is that ordinary mixups are not effective at discriminating numbers that are simple geometric shapes. However, in ICT and MixMatch, the use of mixup for labeled data and unlabeled data is the core of the method. On the other hand, in MixGDA, we can choose either mixup or self-mixup. Self-mixup is a gentler mixup method than mixup, and is also effective for discriminating numbers. As a result, MixGDA achieves better performance than ICT and MixMatch. Another consequence of this idea is that $\delta^{(x+u)}$ for SVHN in Table 1 is set to zero. This indicates that collaborative training is not used for SVHN. The reason for this is that collaborative training randomly mixes up labeled data samples into unlabeled data samples. On the other hand, MixMatch achieves the highest performance for CIFAR-10. MixGDA has severe performance degradation when the number of labeled data is 250. In particular, a large standard deviation of the test error rate indicates that training using MixGDA is not stable in this case. On the other hand, MixMatch maintains stable performance even when the number of labeled data is small. This is a major feature of MixMatch. For CIFAR-100, MixGDA improves the best performance to date by more than 3\%. MixMatch cannot be compared because there are no experimental results. However, MixGDA had to adopt self-mixup instead of mixup for supervised training. From this, it is possible that MixGDA exceed the performance of MixMatch, as in the case of SVHN.\\
\\[-2mm]
In MixGDA, the averaged model always achieves a lower test error rate than the prime model. This shows the usefulness of the averaged model.

\begin{table}[h]
  \centering
\begin{tabular}{l | ccc | cccc | c}
\hline
Datasets 										& \multicolumn{3}{| c |}{\text{SVHN}} 			& \multicolumn{4}{| c |}{\text{CIFAR-10}}		& \text{CIFAR-100}	\\
\hline
Methods/Labels								& 250				& 500				&1000				& 250			&1000				&2000			&4000			&10000 		\\ 	
\hline
\multirow{2}{*}{MT+SNTG \cite{16}}			&4.29				&3.99				&3.86				&---			&18.41				&13.64			&10.93			&---			\\										
												&$\pm$0.23		&$\pm$0.24		&$\pm$0.27		&\, 			&$\pm$0.52		&$\pm$0.32	&$\pm$0.14	&\, 			\\		
\multirow{2}{*}{MT+fast-SWA \cite{1}}		&---				&---				&---				&---			&15.58				&11.02			&9.05			&33.62			\\										
												&\, 				&\, 				&\, 				&\, 			&$\pm$0.12		&$\pm$0.23	&$\pm$0.21	&$\pm$0.54 	\\
\multirow{2}{*}{ICT \cite{25}}					&4.78				&4.23				&3.89				&---			&15.48				&9.26			&7.29			&---			\\										
												&$\pm$0.68		&$\pm$0.15		&$\pm$0.04		&\, 			&$\pm$0.78		&$\pm$0.09	&$\pm$0.02	&--- 			\\	
MixMatch \cite{2}								&3.59				&---				&3.39				&\textbf{14.31}			&---				&---			&\textbf{6.84}		&---			\\	
\hline
\multirow{2}{*}{MixGDA (Averaged model)}	&\textbf{3.32}		&\textbf{3.16}		&\textbf{2.85}				&21.29		&\textbf{10.82}		&\textbf{8.36}		&\textbf{6.87}		&\textbf{30.32}			\\										
				&\textbf{$\pm$0.14}		&\textbf{$\pm$0.14}		&\textbf{$\pm$0.08}		&$\pm$2.13 	&\textbf{$\pm$0.41}		&\textbf{$\pm$0.32}	&\textbf{$\pm$0.13}	&\textbf{$\pm$0.34}	\\										
\multirow{2}{*}{MixGDA (Prime model${}^a$)}	&3.64				&3.59				&3.18				&21.75			&11.33				&8.77			&7.13			&30.43			\\										
												&$\pm$0.16		&$\pm$0.24		&$\pm$0.18		&$\pm$2.42 	&$\pm$0.36		&$\pm$0.24	&$\pm$0.08	&$\pm$0.33	\\		

\hline
\end{tabular}

\vspace{5mm}
  \caption{This table shows the average and standard deviation of test errors. In experiments on MixGDA, labeled data $\dl$ is randomly sampled from training data for each experiment. The test error rate is the result of five experiments. MT stands for Mean Teacher \cite{24}. ${}^a \:$The prime model is a model with weights obtained immediately after training ends. Batch normalization statistics for the prime model are calculated in the same way as for the averaged model.}
  \label{tab2}
\end{table}

%%%%%%%%%%%%% Section 5.2 %%%%%%%%%%%%%%%%%%%%%%%%%%
\subsection{Ablation Study}\label{sec5.2}
MixGDA contains various mechanisms. We confirm the effectiveness of each mechanism by performing experiments that remove specific mechanisms, replace them with alternative mechanisms, or add specific mechanisms.\\
\\[-2mm]
Traditional mixups are not effective for supervised training for SVHN. One way to overcome this is Self-mixup. As shown in Table 3, Self-mixup works effectively in MixGDA. \\
\\[-2mm]
The effect of collaborative training is subtle. As shown in Table 4, for CIFAR-10,collaborative training is more effective with more labeled data. When the number of labeled data decreases to 1000, the effect of collaborative training is not seen. This indicates that the effect of collaborative training decreases as the discrimination accuracy decreases. On the other hand, collaborative training is effective for CIFAR-100 even though it is a dataset with low discrimination accuracy. The elucidation of the mechanism of collaborative training and the improvement of its performance are future tasks.\\
\\[-2mm]
MixGDA is a method that does not require ZCA normalization. As shown in Table 5, the performance of MixGDA deteriorates when ZCA normalization is applied to CIFAR-10. gCCB is a adversarial data augmentation. Therefore, it is interesting to compare gCCB with randomCCB that randomly determines the sign on the right-hand side of (22). As shown in Table 5, the performance of MixGDA is better when gCCB is used. As shown in Table 5, gVAT is effective for SVHN but not for CIFAR-10. Finally, compare Inner with entropy minimization. First, it is important to note that MixGDA uses a loss function called $L_\mathrm{rem}$ to minimize the sum of probabilities other than the principal probabilities. Therefore, even without an Inner, MixGDA has the force to make the probability distribution of unlabeled data samples one-hot. As shown in Table 5, entropy minimization degrades MixGDA performance more than without Inner. It is possible that this phenomenon is due to the incompatibility between entropy minimization and $L_\mathrm{rem}$. In any case, Inner is an effective mechanism in MixGDA.

\begin{table}[H]
  \centering
\begin{tabular}{l | ccc}
\hline
Datasets 										& \multicolumn{3}{c}{\text{SVHN}}      							\\
\hline
Ablation/Labels								& 250					& 500					& 1000		   		\\ 
\hline
MixGDA										& 3.32$\pm$0.14		& 3.16$\pm$0.14		& 2.85$\pm$0.08	\\
MixGDA without Self-mixup$\: {}^\text{a}$ 	& 3.52$\pm$0.16		& 3.26$\pm$0.18		& 3.20$\pm$0.08	\\
\hline
\end{tabular}

\vspace{5mm}
  \caption{This table shows the experimental results on the effectiveness of self-mixup. All values are the mean and standard deviation of the test errors of five experiments. The labeled data $\dl$ differs from experiment to experiment. ${}^a \:$This is the case where neither Self-mixup nor mixup is used in supervised training. Therefore, the labeled data sample obtained by the default data augmentation is mixed up with the $\mathrm{gROI} (u)$ generated from the unlabeled data sample $u$.}
  \label{tab3}
\end{table}

\begin{table}[H]
  \centering
\begin{tabular}{l | cccc | c}
\hline
Datasets 										& \multicolumn{4}{| c |}{\text{CIFAR-10}}      													& \text{CIFAR-100}	\\
\hline
Ablation/Labels								& 250					& 1000					& 2000					& 4000					& 10000	   			\\ 
\hline
MixGDA										& 21.29$\pm$2.13	& 10.82$\pm$0.41	& 8.36$\pm$0.32	  	& 6.87$\pm$0.13 		& 30.32$\pm$0.34	\\
MixGDA without collaborative training	  	& 22.07$\pm$1.88	& 10.82$\pm$0.34	& 8.60$\pm$0.14 		& 7.29$\pm$0.22 		& 32.06$\pm$0.14	\\
\hline
\end{tabular}

\vspace{5mm}
  \caption{This table shows the experimental results on the effectiveness of collaborative training. All values are the mean and standard deviation of the test errors of five experiments. The labeled data $\dl$ differs from experiment to experiment.}
  \label{tab4}
\end{table}

\begin{table}[H]
  \centering
\begin{tabular}{l | cc}
\hline
Datasets 												& \text{SVHN}				& \text{CIFAR-10}      	\\
\hline
Ablation/Labels										& 1000						& 4000					   	\\ 
\hline
MixGDA												& 2.85$\pm$0.08${}^*$	& 6.87$\pm$0.13${}^*$	\\
MixGDA with ZCA	  									& --- 						& 8.37$\pm$0.09			\\
MixGDA with random CCB instead of gCCB 		& 3.22$\pm$0.08			& 7.00$\pm$0.09${}^*$	\\
MixGDA without gVAT								& 3.38$\pm$0.03			& --- 						\\
MixGDA with gVAT									& --- 						& 6.93$\pm$0.15${}^*$	\\
MixGDA without Inner								& 3.00$\pm$0.15			& 7.04$\pm$0.03			\\
MixGDA with entropy minimization instead of Inner & 3.05$\pm$0.04			& 7.38$\pm$0.12			\\
\hline
\end{tabular}

\vspace{5mm}
  \caption{This table shows the results of various ablation studies. The values without ${}^*$ are the average and standard deviation of the test errors of three experiments. The values marked with ${}^*$ are the mean and standard deviation of the test errors of five experiments. The labeled data $\dl$ differs from experiment to experiment.}
  \label{tab5}
\end{table}

%%%%%%%%%%%%% Section 6 %%%%%%%%%%%%%%%%%%%%%%%%%%
\section{Conclusions}\label{sec6}

UDA is a versatile method that can achieve the highest discrimination accuracy at the moment. However, UDA is a computationally expensive method. Our research has begun with the motivation to use the gradient information obtained in the training process to achieve efficient SSL.

UDA has shown that the diversity of data used in consistency regularization (CR) is critical to the success of CR. In this paper, we proposed three types of deterministic data augmentation (Gradient-based Data Augmentation (GDA)) using the pixel value gradient information of the posterior probability distribution. We aimed to realize the data diversity effectively using three types of GDA. In addition, ICT and MixMatch showed that the mixup method for labeled data and unlabeled data is also effective for SSL. We constructed an SSL method named MixGDA by combining GDA and the mixup method. The mixup method used in MixGDA includes  Self-mixup we propose. MixGDA achieves the same performance as MixMatch when CIFAR-10 has 4,000 labeled data. However, when the number of labeled data is 250, the discrimination performance achieved by MixGDA is far less than the discrimination performance achieved by MixMatch. On the other hand, for SVHN, the performance of MixGDA exceeds the performance of MixMatch. For CIFAR-100, MixGDA achieves a discrimination performance that surpasses the conventional highest discrimination performance. MixGDA includes various elemental technologies inspired by past research. Since gVAT, gCCB, and Inner are technologies that can be used separately, it is considered that they can be incorporated into other SSL methods. Also, Self-mixup can be used as a useful data augmentation for datasets where conventional mixup is inappropriate.

Future work is to elucidate the mechanism of collaborative training and improve its performance, and to evaluate MixGDA for Wide-ResNet-28-2.

\bibliographystyle{unsrt}  
%\bibliography{references}  %%% Remove comment to use the external .bib file (using bibtex).
%%% and comment out the ``thebibliography'' section.

%%% Comment out this section when you \bibliography{references} is enabled.

%%%%%%%%%%%%%%%%%%%%%%%%%%%%%%%%%%%%%%%
%%%%%%%%%%%%% Appendix %%%%%%%%%%%%%%%%%%%%%%%%%%
%%%%%%%%%%%%%%%%%%%%%%%%%%%%%%%%%%%%%%%
\vspace{10mm}
\appendix
\section{13-layer CNN used in our experiments}\label{appA}
\begin{table}[H]
\centering
\begin{tabular}{ll} 
\hline
Layer & Hyperparameters \\ 
\hline
Convolution${}^\text{a}$ + BN + Leaky ReLU (0.1) & 128 filters, $3 \times 3$ \\
Convolution${}^\text{a}$ + BN + Leaky ReLU (0.1) & 128 filters, $3 \times 3$ \\
Convolution${}^\text{a}$ + BN + Leaky ReLU (0.1) & 128 filters, $3 \times 3$ \\
Pooling + Dropout ($p = 0.5$) & Maxpool $2 \times 2$, stride 2 \\
Convolution${}^\text{a}$ + BN + Leaky ReLU (0.1) & 256 filters, $3 \times 3$ \\
Convolution${}^\text{a}$ + BN + Leaky ReLU (0.1) & 256 filters, $3 \times 3$ \\
Convolution${}^\text{a}$ + BN + Leaky ReLU (0.1) & 256 filters, $3 \times 3$ \\
Pooling + Dropout ($p = 0.5$) & Maxpool $2 \times 2$, stride 2 \\
Convolution${}^\text{a}$ + BN + Leaky ReLU (0.1) & 512 filters, $3 \times 3$ \\
Convolution${}^\text{a}$ + BN + Leaky ReLU (0.1) & 256 filters, $1 \times 1$ \\
Convolution${}^\text{a}$ + BN + Leaky ReLU (0.1) & 128 filters, $1 \times 1$ \\
Pooling & Global average pooling ($6 \times 6 \rightarrow 1 \times 1$) \\
Fully connected${}^\text{a}$ + BN${}^\text{b}$ + Softmax & $128  \rightarrow  10$${}^\text{c}$ \\
\hline               
\end{tabular}
\vspace{5mm}
  \caption{The convolutional network architecture used in our experiments. BN refers to batch normalization using the mean and standard deviation on each minibatch. ${}^\text{a}$ For CIFAR-10 and CIFAR-100, apply weight normalization [T. Salimans (2016)]. ${}^\text{b}$ Not applied on CIFAR-10 and CIFAR-100 experiments. ${}^\text{c}$ In the case of CIFAR-100, use $128 \rightarrow 100$.}
  \label{tab6}
\end{table}

\end{document}